\documentclass[runningheads]{llncs}
\usepackage{graphicx}
\usepackage[utf8]{inputenc}
\usepackage[english]{babel}
\usepackage{comment}
\usepackage{subfig}
\usepackage[backend=bibtex]{biblatex}
\graphicspath{{./figs/}}
\newcommand\drafters{12}
\newcommand\circuitsPerDrafter{12}
\newcommand\imagesPerCircuit{8}
\newcommand\classes{45}
\newcommand\circuits{\the\numexpr\drafters*\circuitsPerDrafter\relax}
\newcommand\images{\the\numexpr\drafters*\circuitsPerDrafter*\imagesPerCircuit\relax}
\newcommand\minWidthImage{864}
\newcommand\maxWidthImage{4\,160}
\newcommand\minHeightImage{396}
\newcommand\maxHeightImage{4\,032}
\newcommand\annotationCount{48\,563}
\newcommand\junctionCount{18\,229}
\newcommand\textCount{14\,231}
\newcommand\crossoverCount{1\,396}

\begin{document}

\title{A Public Ground-Truth Dataset for Handwritten Circuit Diagram Images}
\titlerunning{Public Handwritten Circuit Images Dataset}

\author{Felix Thoma\inst{1\and 2} \and Johannes Bayer\inst{1 \and 2} \and Yakun Li\inst{1}}
\authorrunning{Thoma et al.}
\institute{Smart Data \& Knowledge Services Department, DFKI GmbH\\ Kaiserslautern, Germany\\
\email{\{felix.thoma,johannes.bayer,yakun.li\}@dfki.de} \and
Computer Science Department, TU Kaiserslautern\\ Kaiserslautern, Germany}

\maketitle

\begin{abstract}
The development of digitization methods for line drawings – especially in the area of electrical engineering – relies on the availability of publicly available training and evaluation data. This paper presents such an image set along with annotations. The dataset consists of $\images$ images of $\circuits$ circuits by $\drafters$ drafters and $\annotationCount$ annotations. Each of these images depicts an electrical circuit diagram, taken by consumer grade cameras under varying lighting conditions and perspectives. A variety of different pencil types and surface materials has been used. For each image, all individual electrical components are annotated with bounding boxes and one out of $\classes$ class labels. In order to simplify a graph extraction process, different helper symbols like junction points and crossovers are introduced, while texts are annotated as well. The geometric and taxonomic problems arising from this task as well as the classes themselves and statistics of their appearances are stated. The performance of a standard Faster RCNN on the dataset is provided as an object detection baseline.
\end{abstract}

\section{Introduction}
\label{sec:intro}
The sketch of a circuit diagram is a common and important step in the digital system design process. Even though the sketch gives the designer a lot of freedom to revise, it is still time consuming to transform the sketch into a formal digital representation for existing tools to simulate the designed circuit. There are mainly two approaches which have been applied to address the transformation from sketch into a useful format to be used by existing simulation tools. The first approach allows the user to draw the diagram in an on-line fashion on a screen and tries to recognize the sketch by interpreting the pen strokes of user, like shown in \cite{echalk}, \cite{sketchread} and \cite{logisketch}. The second, off-line approach assumes that the diagrams are drawn with a pen on paper, from which an image is captured afterwards. This approach relies on machine learning and computer vision algorithms, like shown in \cite{Edwards2000MachineRO}, \cite{Moetesum2018SegmentationAR}, \cite{Patare2016HanddrawnDL} and \cite{sketic}, to process the images. It is important to collect large and high quality training data for the machine learning and computer vision approach, especially when deep learning algorithms are applied.

\begin{figure}
  \subfloat[Full Image\label{fig:sample_overview}]{
    \includegraphics[width=0.49\textwidth]{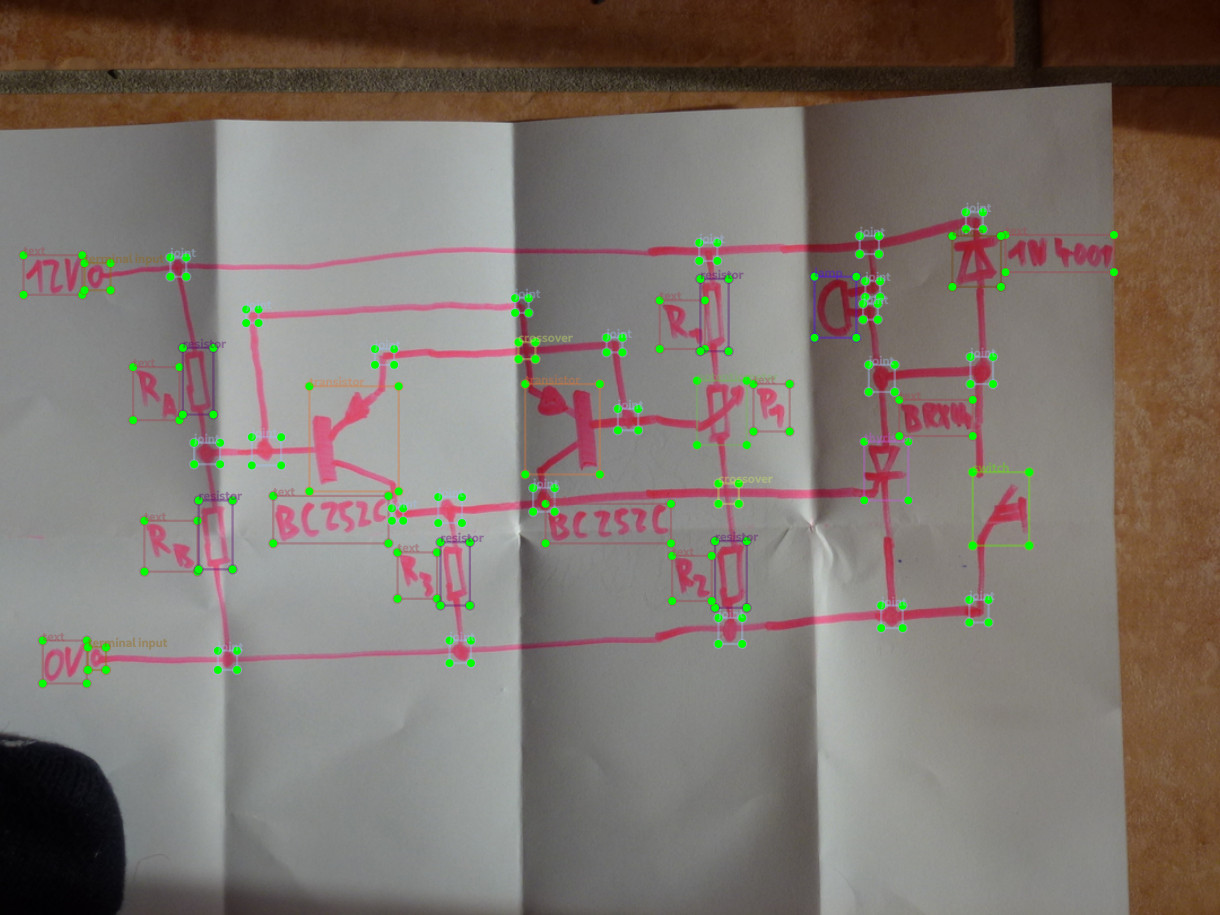}
  }
  \hfill
  \subfloat[Detail showing Junction, Crossover and Text Annotations\label{fig:sample_detail}]{
    \includegraphics[width=0.49\textwidth]{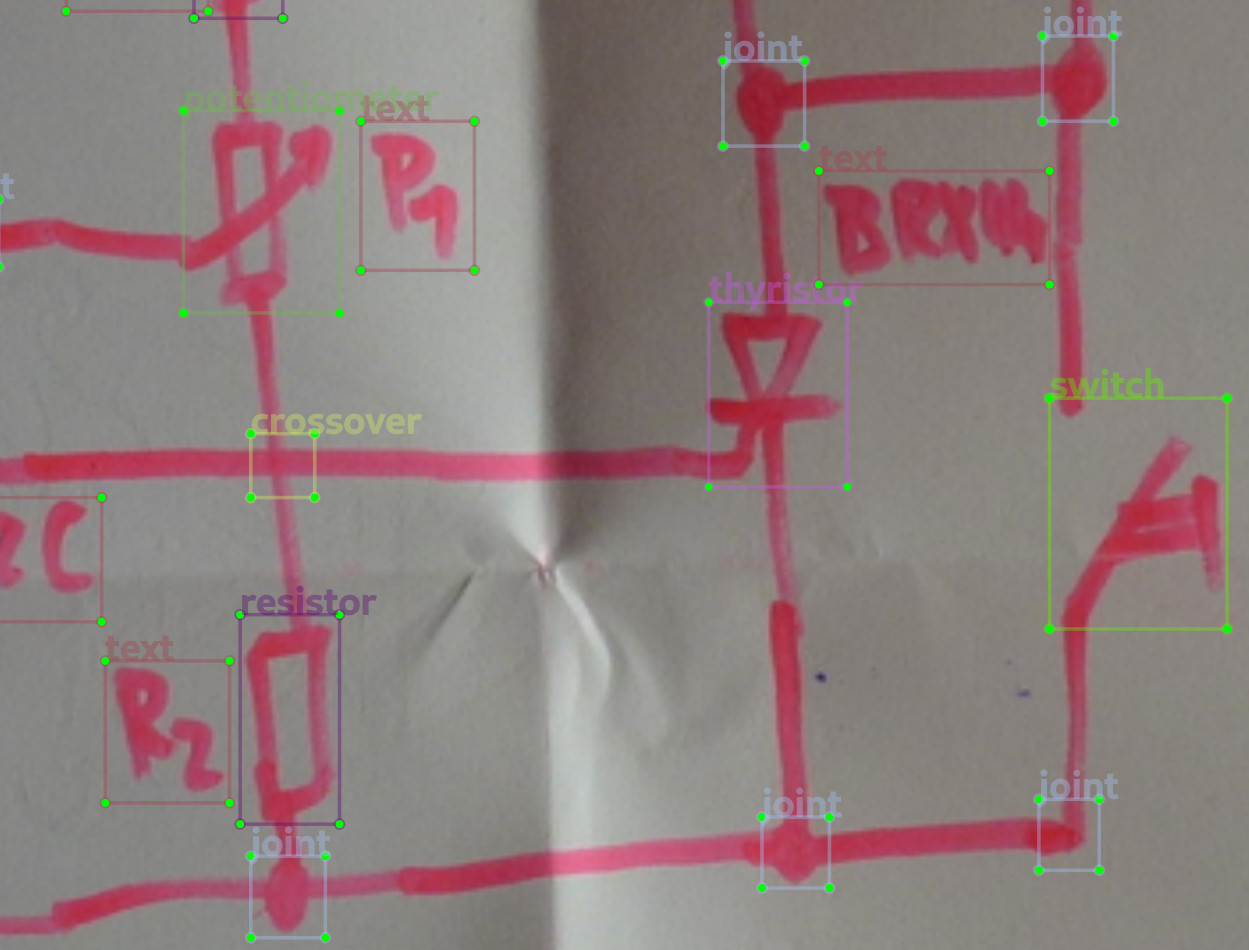}
  }
  \caption{Dataset Sample with Annotations}
  \label{fig:sample}
\end{figure}

However, most of this literature\cite{Edwards2000MachineRO}\cite{Moetesum2018SegmentationAR}\cite{Patare2016HanddrawnDL}\cite{sketic} exposes only little of the data it relies on publicly and to the best of the  knowledge of this paper's authors, there is no publicly available comparable dataset.

The dataset described in this paper mainly addresses this issue and will be made publicly available\footnote[1]{\url{https://git.opendfki.de/circuitgraph-public/cghd/}} under the MIT Licence\footnote[2]{\url{https://opensource.org/licenses/MIT}}. Previous datasets described in literature will be summarized in section \ref{sec:relatedwork}. Afterwards, the characteristics of the hand-drawn circuit diagrams images contained in the dataset will be stated in section \ref{sec:images}. In the end, section \ref{sec:baseline} will present basic recognition results using a Faster-RCNN \cite{fasterrcnn} on the presented dataset.

\section{Related Work}
\label{sec:relatedwork}
The current research papers for the off-line approach mentioned in section \ref{sec:intro} mainly focus on the algorithms to recognize the sketches and their datasets are not publicly accessible. While \cite{sketic} presented a comprehensive system to recognize hand drawn sketches and transform the recognition results to Verilog code, there was no description of the concrete number of images and the variety of symbols. In \cite{Patare2016HanddrawnDL}, the author proposed to use Fourier Descriptors to get feature representation of hands drawn circuit components and then use SVM to classify them. They used 10 hand drawn circuit sketches as test data, but only one circuit sketch was shown as an example. The author of \cite{Edwards2000MachineRO} gave information of the number of nodes and components included in their hand drawn sketches but without the exact number of sketches. \cite{Moetesum2018SegmentationAR} mentioned that they did their evaluation on 100 hand drawn circuit diagrams, however the quantity is relatively small compared with the dataset presented in the current paper.

\section{Images}
\label{sec:images}
Every image of the dataset contains a hand-drawn electrical circuit (see fig.~\ref{fig:sample}). Every of the $\drafters$ drafters was instructed to draw $\circuitsPerDrafter$ circuits, each of them $2$ times, where it was allowed the drawers to alter the circuit geometrically. Each of the drafts was photographed $4$ times, resulting in an overall image count of $\images$. Taking multiple images of the same drawing was based on the following considerations:

\begin{itemize}
 \item The dataset can conveniently be used for style transfer learning
 \item Since different images of the same drawing contain the same annotation items, they can be used for automatic verification of the labeling process
 \item The cost of creating a single sample is reduced
 \item Different real-world image captures are a realistic and rich replacement for artificial data augmentation
 \item Different samples of the same drawing must be carefully annotated to ensure the object detection task is well-defined (no ambiguities)
 \item enforcing diverse capturing conditions aim to support camera feeds
\end{itemize}

On a semantic level the user can draw a similar circuit using different symbols, for example to create a voltage source a DC source can be used, a Vcc together with GND could be used, or only the Terminals could be drawn with textual description of the meaning of it.

\begin{comment}
\begin{figure}
  \subfloat[D1-P1\label{fig:sample_overview}]{
    \includegraphics[width=0.23\textwidth]{C54_D1_P1}
  }
  \hfill
  \subfloat[D1-P2\label{fig:sample_detail}]{
    \includegraphics[width=0.23\textwidth]{C54_D1_P2}
  }
  \hfill
  \subfloat[D1-P3\label{fig:sample_overview}]{
    \includegraphics[width=0.23\textwidth]{C54_D1_P3}
  }
  \hfill
  \subfloat[D1-P4\label{fig:sample_detail}]{
    \includegraphics[width=0.23\textwidth]{C54_D1_P4}
  }
  
  \subfloat[D2-P1\label{fig:sample_overview}]{
    \includegraphics[width=0.23\textwidth]{C54_D2_P1}
  }
  \hfill
  \subfloat[D2-P2\label{fig:sample_detail}]{
    \includegraphics[width=0.23\textwidth]{C54_D2_P2}
  }
  \hfill
  \subfloat[D2-P3\label{fig:sample_overview}]{
    \includegraphics[width=0.23\textwidth]{C54_D2_P3}
  }
  \hfill
  \subfloat[D2-P4\label{fig:sample_detail}]{
    \includegraphics[width=0.235\textwidth]{C54_D2_P4}
  }  
  
  \caption{Images Samples Drawn from a Single Circuit. The two drawings not only differ in background and pencil type, but also in their layout. Despite varying degrees of disturbances, all components remain basically recognize.}
  \label{fig:sample_c}
\end{figure}
\end{comment}

\subsection{Drawing Surfaces and Instruments}
The depicted surface types include: plain paper of various colors, ruled paper, squared paper, coordinate paper, glas, whiteboard, cardboard and aluminum foil. Pens, pencils, permanent markers and whiteboard markers of different colors have been utilized for drawing. A few drawings contain multiple colors, or contain text that has been highlighted by a dedicated color. Buckling, kinking, bending, spots, transillumination, and paper cracks can be found as distortions on the samples. Some samples have been treated by corrective means, others have been created using a ruler as a drawing aid rather than free-hand sketching.

\subsection{Capturing}
All images have been captured by consumer cameras, no artifical image distortion has been applied. Some images have been cropped and company logos have been obliterated. Camera position and spatial orientation alternations as well as steam have been used in various degrees. That way, naturally observable distortions like (motion) blur and light reflexes of surfaces as well as ink are captured in the dataset. Despite these disturbances no major occlusions are present. More precisely, circuit symbols and their connections are recognizable by humans and the circuits are oriented right side up. Often, circuit drawings are surrounded by common (office environment) objects. The sample image sizes range between $\minWidthImage$ by $\maxWidthImage$ and $\minHeightImage$ by $\maxHeightImage$ and are stored as \texttt{JPEG}.

\section{Annotations}
Bounding box annotations stored in the PascalVOC\cite{pascalvoc} format have been chosen for symbols and texts as a trade-off between effort in ground truth generation and completeness in capturing the available information from the images. The downsides of this decision are the lack of rotation information of the symbols as well as the lack of wire layout capturing  and will be addressed by the auxiliary classes below. The annotations have been created manually using labelIMG\cite{labelimg}.

\subsection{Auxilary Classes}
In order to address the lack of wire-line annotation and to simplify the extraction of a circuit graph, auxiliary classes are introduced:

\begin{itemize}
 \item \textbf{junction} represents all connections or kink-points of wire edges. Since wire corners and junctions share the same semantic from the electrical graph point of view, they share the same label.
 
 \item \textbf{crossover} represents the optical crossing of a pair of lines that is not physically connected.

 \item \textbf{terminal} denotes an 'open' wire ending, i.e. a circuit input or output.
 
 \item \textbf{text} encloses a line of text, usually physical quantities or terminal or component name.
\end{itemize}

Some symbols (like potentiometers and capacitors) have outgoing lines or corners (like transistors or switches) which are considered part of their definition. Hence, these line (parts) are also included in the bounding boxes and junction annotations within them are neglected. Likewise, text annotations related to symbols are only omitted if they are part of the symbol definition (like an 'M' in the motor symbol). If they are however explaining component parts (like integrated circuit pins) or the component itself (like integrated circuit type labels), they are annotated (even if they are located inside the symbol).

\begin{figure}[h]
  \centering
  \includegraphics[width=0.8\linewidth]{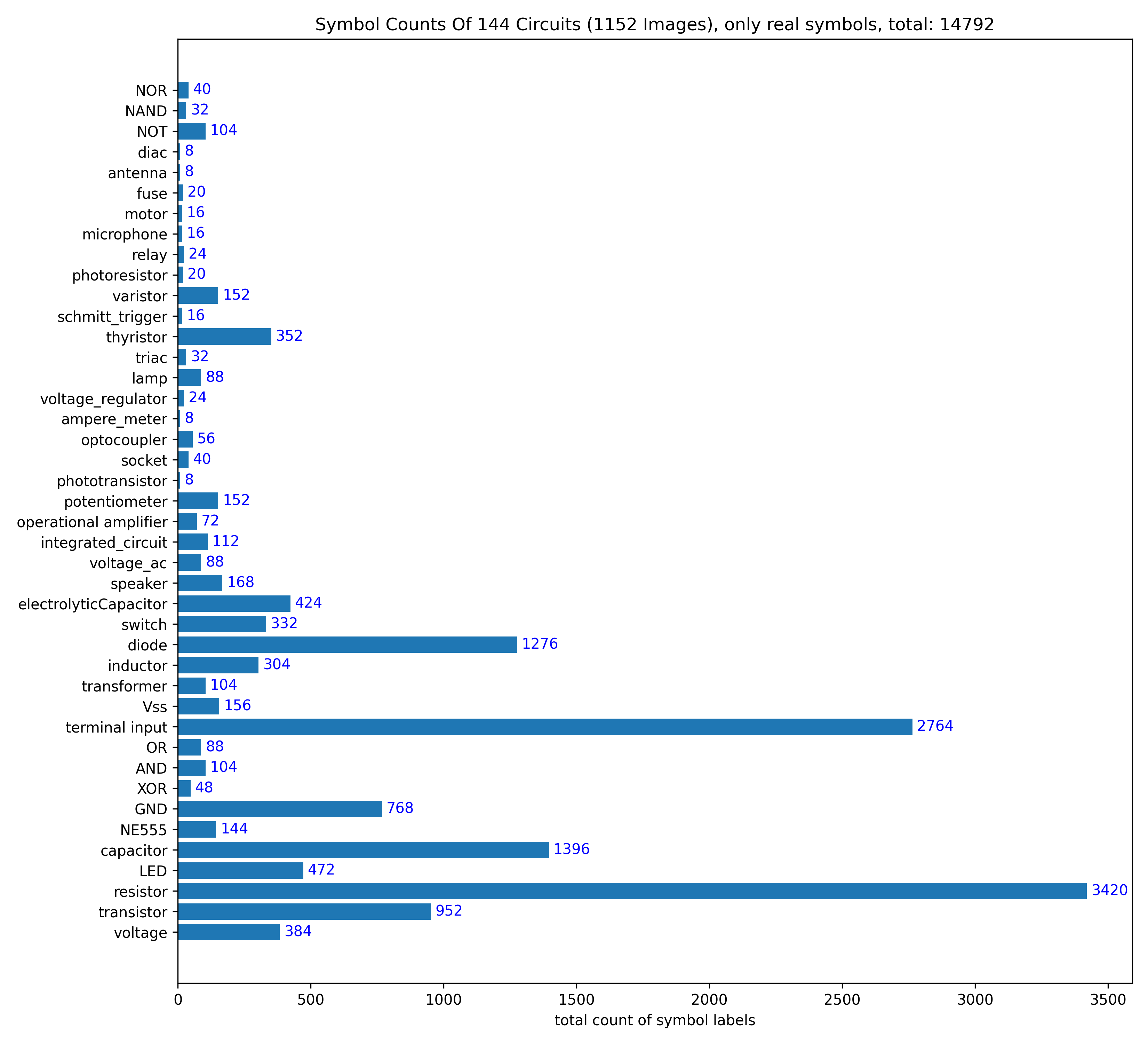}
  \caption{Total count of label instances in the full dataset. Junctions, crossover and text are not considered, due to their dominance in the dataset.}
  \label{fig:statistics}
\end{figure}

\subsection{Symbol Classes}
The dataset contains both IEE/ANSI as well as IEC/DIN symbols. Individual circuit drawings are allowed to contain both variants as well as a mix of analogue and digital symbols. Some drafters also used non standard-conform symbols, for example a speaker explicitly containing a coil to show its exact implementation.

The classes have been chosen to represent \textit{electrically} connectable components. For example, if there is a full-wave rectifier depicted as four diodes, each of the diodes is annotated individually. Likewise, a relay label is used as an annotation rather than a switch and a inductor (or the two inductors of a transformer), since the components are coupled \textit{inductively}. However, if the individual parts are separated so that other components are in between, an individual labeling is used as a makeshift. To simplify annotation, some of the electrical taxonomy is neglected. For example, all bipolar transistors, Mosfets and IGBTs are labeled \textit{transistors}.

\subsection{Geometry}
Generally, the bounding box annotations have been chosen to completely capture the symbol of interest, while allowing for a margin that includes the surrounding space. Often, there is an overlap between tightly placed symbols. The task of fully capturing all parts of the symbol was not well-defined in the context of curvy straight lines connected by rounded edge corners.

\section{Statistics}
\begin{figure}[h]
  \subfloat[Symbol Count Histogram \label{fig:histogram}]{
    \includegraphics[width=0.49\textwidth]{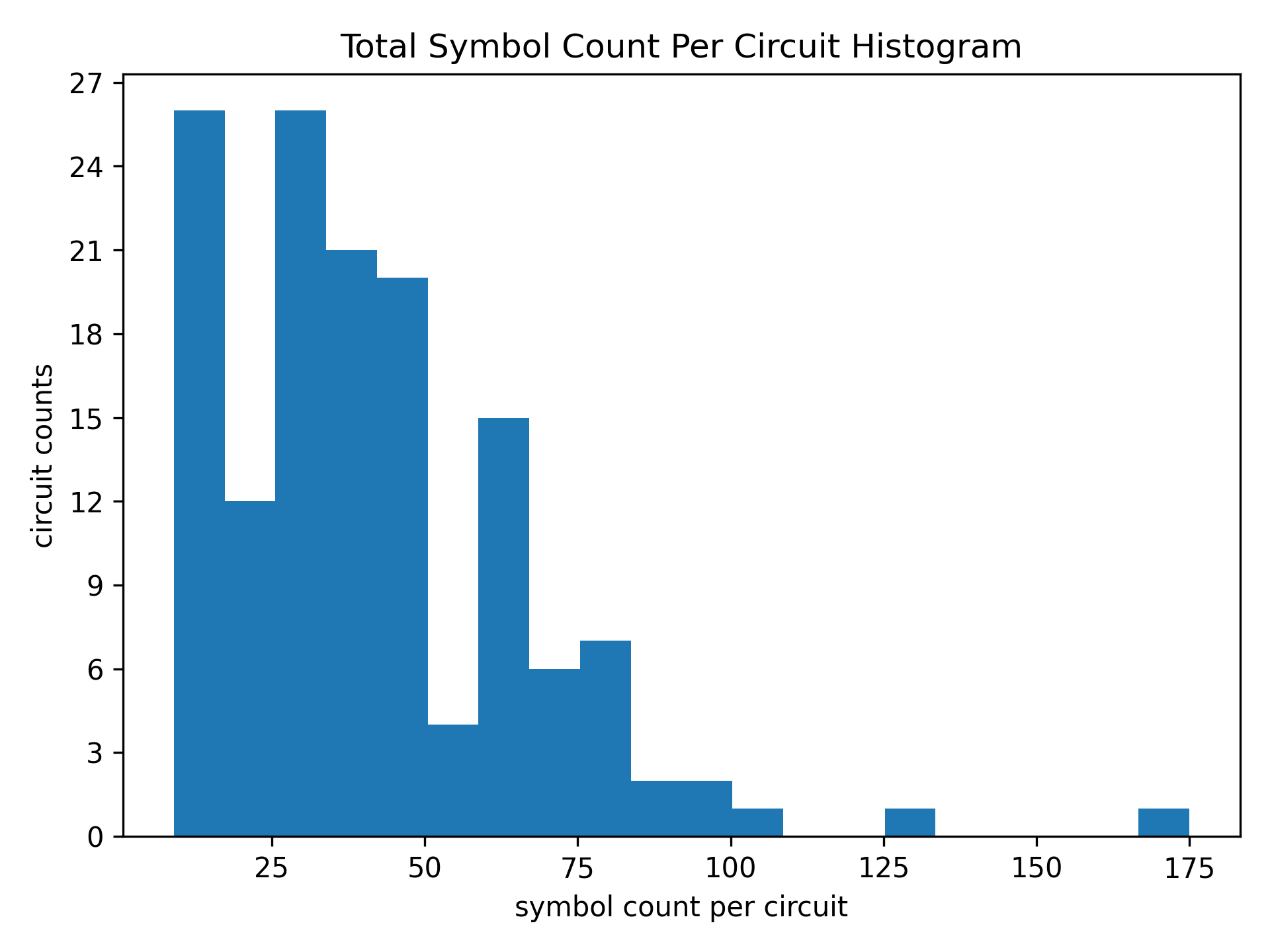}
  }
  \hfill
  \subfloat[Symbol Distribution \label{fig:distribution}]{
    \includegraphics[width=0.49\textwidth]{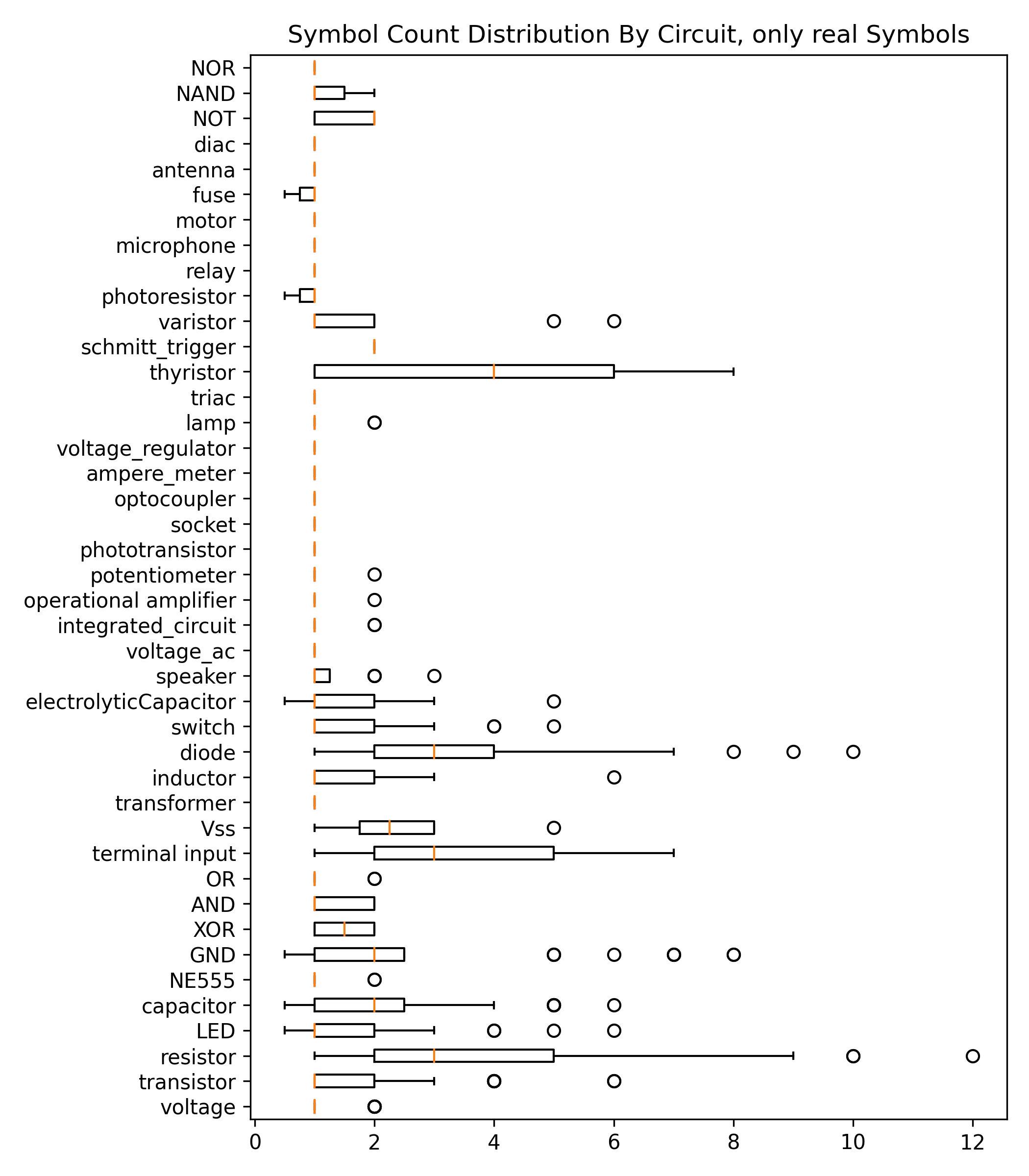}
  }
  \caption{a) Total symbol count per circuit. Histogram is counting the occurences in 20 evenly spaced bins.\\
  b) Distribution of symbol count per circuit that the symbol is part of. Junctions, crossover and text are not considered, due to their dominance in the dataset.}
  \label{fig:percircuit}
\end{figure}

The dataset has a total amount of $\annotationCount$ annotations among $\classes$ classes. The junction class occurs $\junctionCount$ times, the text class $\textCount$ times and the crossover class $\crossoverCount$ times. All other electrical symbol counts are listed in detail in fig.~\ref{fig:statistics}. The frequency of symbol occurences per circuit are displayed in fig.~\ref{fig:percircuit}.

\section{Baseline Performance}
\label{sec:baseline}
A suggested division of the dataset into subsets for training ($125$ circuits), validation ($7$ circuits) and testing ($12$ circuits, all from the same drafter), is provided so that images of every circuit (and therefore drawing) are exclusively assigned to one of the sets only.

In order to establish a baseline performance on the dataset, a Faster RCNN\cite{fasterrcnn} with a ResNet152\cite{resnet} backbone was applied using the Torchvision\cite{torchvision} implementation, a learning rate of $0.007$ and a batch size of $8$, yielding an mAP of 52\%.

\section*{Acknowledgement}
The authors coardially thank Thilo P\"utz, Anshu Garg, Marcus Hoffmann, Michael Kussel, Shahroz Malik, Syed Rahman, Mina Karami Zadeh, Muhammad Nabeel Asim and all other drafters who contributed to the dataset. This research was funded by the german Bundesministerium f\"ur Bildung und Forschung (Project SensAI, grant no. 01IW20007).

\printbibliography

\end{document}